\documentclass[journal]{IEEEtran} 
\pdfoutput=1
\usepackage[utf8]{inputenc}         
\usepackage[T1]{fontenc}            
\usepackage{hyperref}               
\usepackage{xurl}
\usepackage{nicefrac}               
\usepackage{microtype}              
\usepackage{amsfonts}               
\usepackage{amsmath}                
\usepackage[dvipsnames]{xcolor}     
\usepackage{float}                  
\usepackage{graphicx}               
\graphicspath{{images/}}
\DeclareGraphicsExtensions{.pdf,.PDF,.jpg,.JPG,.jpeg,.JPEG,.png,.PNG}
\usepackage[caption=false]{subfig}  
\usepackage{booktabs}               
\usepackage{multicol}               
\usepackage{multirow}               
\usepackage{booktabs}               
\usepackage{array}                  
\newcolumntype{C}[1]{>{\centering\arraybackslash}p{#1}}

\usepackage[nolist]{acronym}        
\usepackage{verbatim}               
\usepackage{comment}                

\newcommand{\white}[1]{{\color{white}#1}}

\begin{document}

\title{Labeling, Cutting, Grouping: an Efficient Text Line Segmentation Method for Medieval Manuscripts}

\author{
    \IEEEauthorblockN{
        \textbf{Michele~Alberti}\IEEEauthorrefmark{1}\IEEEauthorrefmark{2}, \and
        \textbf{Lars~V\"ogtlin}\IEEEauthorrefmark{1}\IEEEauthorrefmark{2}, \and
        \thanks{\IEEEauthorrefmark{1} Both authors contributed equally to this work.}
        \textbf{Vinaychandran~Pondenkandath}\IEEEauthorrefmark{2}, \and
        \textbf{Mathias~Seuret}\IEEEauthorrefmark{2}\IEEEauthorrefmark{3}, \and \\
        \textbf{Rolf~Ingold}\IEEEauthorrefmark{2}, \and
        \textbf{Marcus~Liwicki}\IEEEauthorrefmark{2}\IEEEauthorrefmark{4}
    }\\
    \vspace{0.2cm}
    \IEEEauthorblockA{
        \IEEEauthorrefmark{2}%
        \textit{Document Image and Voice Analysis Group (DIVA)} \\
        University of Fribourg, Switzerland\\
        \{firstname\}.\{lastname\}@unifr.ch \\
        \vspace{0.15cm}
        \IEEEauthorrefmark{3}%
        \textit{Pattern Recognition Lab} \\
        Friedrich-Alexander-Universit\"at Erlangen-N\"urnberg, Erlangen, Germany\\
        mathias.seuret@fau.de\\
        \vspace{0.15cm}
        \IEEEauthorrefmark{4}%
        \textit{Machine Learning Group} \\
        Lule{\aa} University of Technology, Sweden\\
        marcus.liwicki@ltu.se\\
    }
}

\markboth{}{First author et al. : Title}

\maketitle

\thispagestyle{empty}

\begin{acronym}[Bash]
    \acro{AI}{Artificial Intelligence}
    \acro{RL}{Reinforcement Learning}
    \acro{MLP}{Multilayer Perceptron}
    \acro{ANN}{Artificial Neural Network}
    
    \acro{OCR}{Optical Character Recognition}
    \acro{DIA}{Document Image Analysis}
    
    \acro{CC}{Connected Component}
    \acro{MST}{Minimum Spanning Tree}

\end{acronym}

\begin{abstract}


This paper introduces a new way for text-line extraction by integrating deep-learning based pre-classification and state-of-the-art segmentation methods.
Text-line extraction in complex handwritten documents poses a significant challenge, even to the most modern computer vision algorithms.
Historical manuscripts are a particularly hard class of documents as they present several forms of noise, such as degradation, bleed-through, interlinear glosses, and elaborated scripts. 
In this work, we propose a novel method which uses semantic segmentation at pixel level as intermediate task, followed by a text-line extraction step.
We measured the performance of our method on a recent dataset of challenging medieval manuscripts and surpassed state-of-the-art results by reducing the error by 80.7\%.
Furthermore, we demonstrate the effectiveness of our approach on various other datasets written in different scripts.
Hence, our contribution is two-fold.
First, we demonstrate that semantic pixel segmentation can be used as strong denoising pre-processing step before performing text line extraction.
Second, we introduce a novel, simple and robust algorithm that leverages the high-quality semantic segmentation to achieve a text-line extraction performance of 99.42\% line IU on a challenging dataset.
\end{abstract}

\section{Introduction}
\label{toc:introduction}

\global\csname @topnum\endcsname 0
\global\csname @botnum\endcsname 0

\begin{figure}[t]
\centering

\subfloat[RGB domain]{
\includegraphics[width=.29\columnwidth]{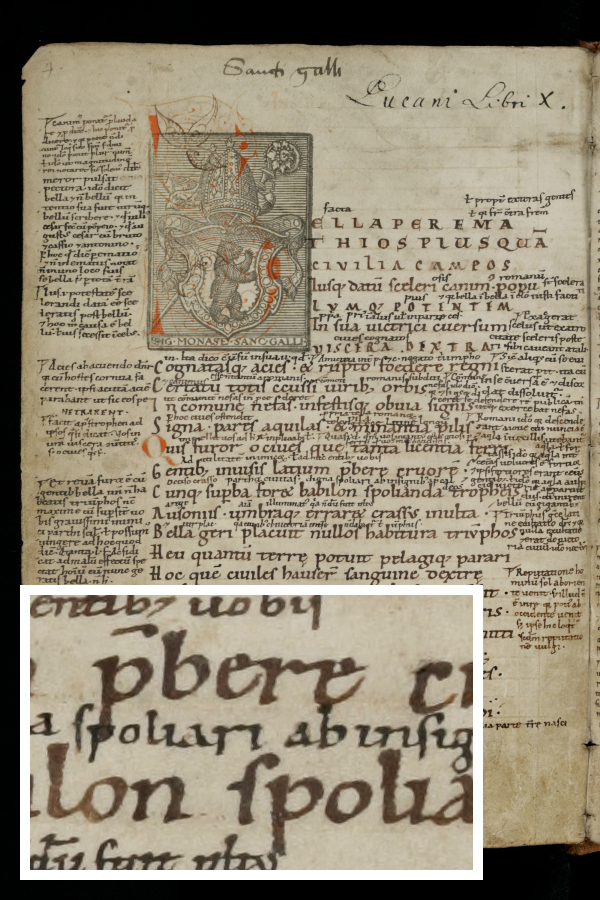}\label{fig:task_rgb}}
\hfil
\subfloat[Pixel segmentation]{
\includegraphics[width=.29\columnwidth]{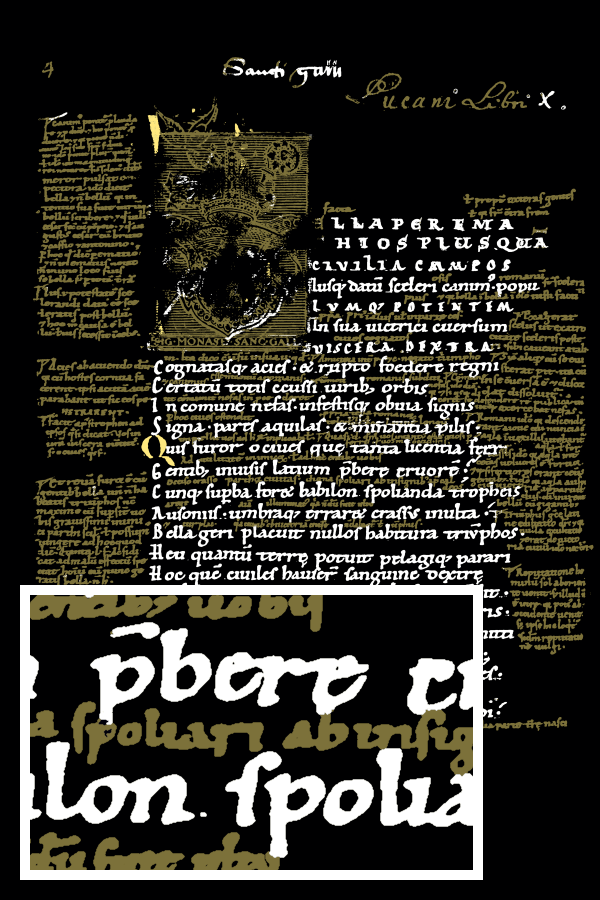}\label{fig:task_semantic}}
\hfil
\subfloat[Line segmentation]{
\includegraphics[width=.29\columnwidth]{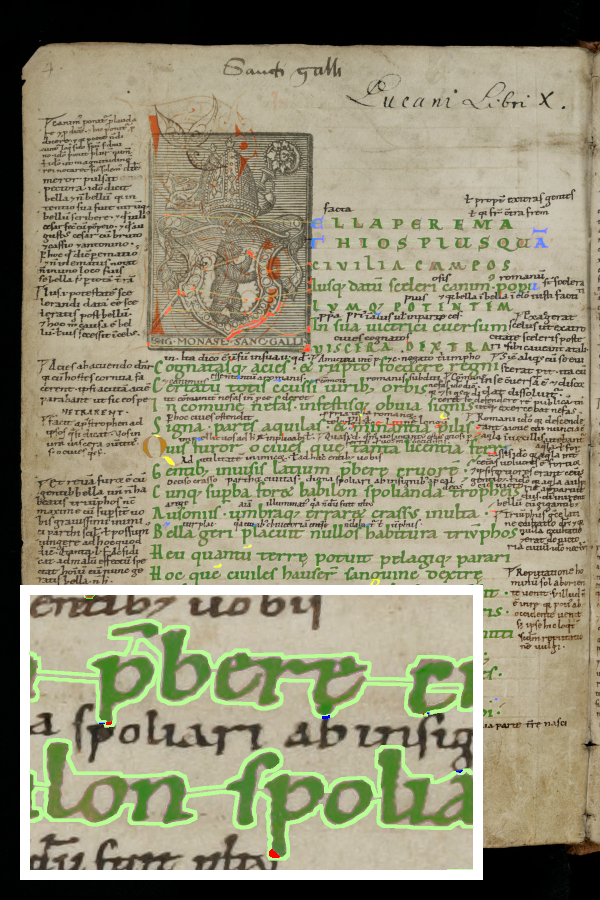}\label{fig:task_lines}}

\caption{%
Stages in the pipeline of our method. 
First, we go from RGB domain (a) to the pixel-label domain (b) by performing semantic segmentation at pixel-level.
Then, we perform text-line segmentation with our novel approach and produce the enclosing polygons for each text-line (c).
The colors in these visualizations are only representational: in (b) text pixels are white and other classes are different shades of brown, in (c) light green is used to mark the output polygons of our algorithm and dark green to denote the pixels enclosed in such polygons.
} 
\label{fig:task}
\end{figure}
Text-line segmentation is a crucial part of document image processing and remains mainly unsolved, especially in documents with complex layouts \cite{sivatanabe2000layout,diem2017cbad,simistira2017icdar2017}.
While nearly-perfect commercial tools exist for scanned modern texts and historical printed texts with simple layouts \cite{papadopoulos2013impact}, other documents (handwritten documents, ornamented documents, and advertisements) pose more challenges.
Especially historical manuscripts, which are the main focus of this paper, suffer from degradation, contain ornaments and decorations, contain varying font sizes and scripts, and often have interlinear and marginal glosses (see Section~\ref{toc:experimental_setting}).

Due to these challenges, many different approaches have been used (see Section~\ref{toc:related_Work}).
However, the use of pre-classification with deep learning based approaches has not been investigated so far\footnote{Note that at ICDAR 2017 Seuret \textit{et al.} \cite{simistira2017icdar2017} have used a preliminary version of our proposed method for demonstrating the possibility of classification as proxy task, however, this paper presents a more elaborated and robust method.}.


In this paper we propose a novel method which leverages high-quality semantic segmentation at pixel-level as a denoising tool, followed by an efficient and robust algorithm to segment the lines into tight polygons.
The first key contribution is showing how semantic segmentation at pixel-level can be used as strong denoising step used as pre-processing before performing text-line extraction.
The second key contribution is introducing a novel simple algorithm that leverages this data preparation, achieving promising results on a challenging dataset of medieval manuscripts.
Finally, we open-source\footnote{\url{https://github.com/DIVA-DIA/Text-Line-Segmentation-Method-for-Medieval-Manuscripts}} our code such that other researchers can benefit from it.

\begin{figure*}[t] 
\centering

\subfloat[CB55, p.25r]{
\includegraphics[width=\columnwidth]{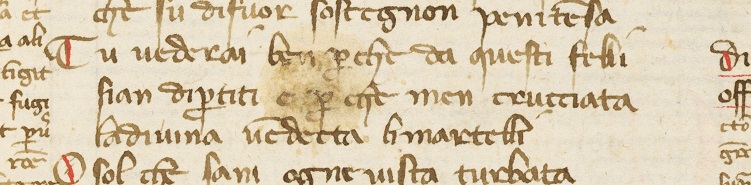}\label{fig:dataset_cb55_A}}
\hfill
\subfloat[CB55, p.32v]{
\includegraphics[width=\columnwidth]{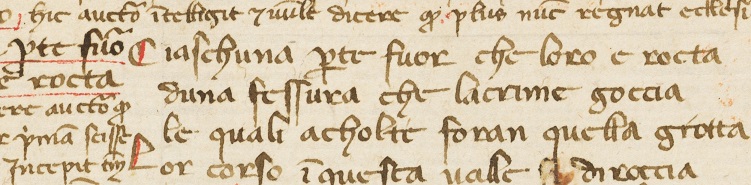}\label{fig:dataset_cb55_B}}

\vfill

\subfloat[CSG18, p.107]{
\includegraphics[width=\columnwidth]{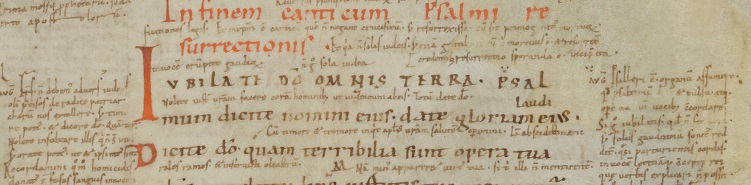}\label{fig:dataset_csg18_A}}
\hfill
\subfloat[CSG18, p.48]{
\includegraphics[width=\columnwidth]{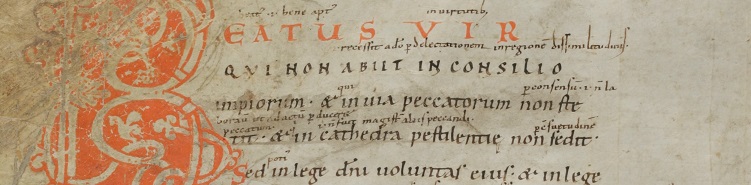}\label{fig:dataset_csg18_B}}

\vfill

\subfloat[CSG863, p.4]{
\includegraphics[width=\columnwidth]{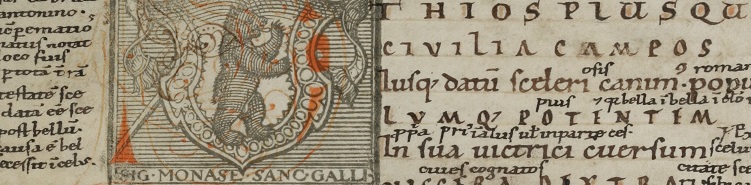}\label{fig:dataset_csg863_A}}
\hfill
\subfloat[CSG863, p.131]{
\includegraphics[width=\columnwidth]{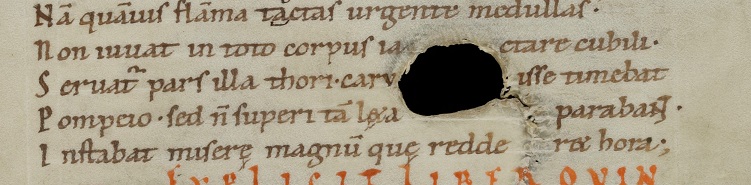}\label{fig:dataset_csg863_B}}

\caption{
Samples of pages of the three medieval manuscripts in DIVA-HisDB, where it is possible to observe some of the multiple features which make this dataset challenging.
For example, one can see paper degradation (a), red strokes in lettrines (b), small difference in color and style between text and comments (b), red text and different lettrine sizes (c), colored decorations, titles and interlinear glosses (d), black decorations and different font sizes (e), critical degradation i.e. part of the page missing (f).
}
\label{fig:dataset}
\end{figure*}

\section{Related Work}
\label{toc:related_Work}

A well-known text-line segmentation method presented by Wong \textit{et al.}~\cite{wong1982document} is based on image smearing, which consists of binarizing a document image, and expanding the foreground horizontally and/or vertically to make the connected components merge.
With horizontal smearing, it can detect text-lines, and with both horizontal and vertical smearing, it can detect text blocks.
Simple to implement, this method, however, suffers from two main flaws when applied to medieval manuscripts.
First, a good binarisation method is needed.
Second, it fails if descenders of a line touch ascenders from the next line.

Another more frequently used approach for text-line detection is to use projection profiles, i.e., to sum the values of the pixels of a row, either in grayscale or a binary version of the image~\cite{razak2008off}.
The positions of the peaks and valleys of the projections correspond to the position of the text lines and spaces between text-lines, as long as several conditions are met: the projection must be roughly parallel to the lines, the lines must be straight, and the interlinear space must be sufficiently large.
This approach was improved by Shapiro \textit{et al.}~\cite{shapiro1993handwritten}, who showed that Hough transforms can be used for modifying the page in such a way that the text-lines are horizontal, and thus easier to separate using projection profiles.
This tends to work very well on modern printed or typed documents~\cite{antonacopoulos2004document}, however, exhibit significantly lower performance on medieval handwritten documents, especially in the presence of marginal or interlinear glosses, or of paratextual elements.

A significantly more efficient approach for text segmentation of historical document images is the use of seam carving~\cite{asi2011text}.
Seam carving consists of computing seams that attempt to separate text-lines; cutting ascenders and descenders if necessary.
This is done by computing an energy map such that the seam stays away as far as possible from the text-lines, and cuts the text (when needed) in the shortest way.

The performance of both projection profiles and seam carving methods can be improved for documents with regular interlinear space, by taking into account the frequency of appearance of text-lines~\cite{seuret2017robust}.
This allows detecting short text-lines, such as a single word at the end of a paragraph, which would have otherwise been missed.

A completely different way is to treat the connected components as graphs.
Garz \textit{et al.} developed a method based on interest point clustering~\cite{garz2012binarization}.
The interest points are computed using Difference of Gaussian~\cite{lowe2004distinctive}, and high-density regions of such points are identified as words.
Seam carving is used to separate these clusters when ascenders or descenders connect them.
Finally, clusters are concatenated with their closest neighbours with regard to the text orientation of the document to produce text-lines.
Pastor~\textit{et al.}~\cite{pastor2013combining} extended this method by aggregating with combinatorial algorithms interest points, such as upper and lower text baselines, computed by a deep neural network.
This offers the advantage to focus on the text, and therefore ignores degradations or drawings.

The closest work related to our method are multi-step methods, presented by Pastor~\textit{et al.}~\cite{pastor2016complete} and Gruüning~\textit{et al.}~\cite{gruuening2017robust}.
The former employs a multi-stage deep learning approach to detect text regions followed by watershed-transform as post-processing step.
The latter, performs an unsupervised clustering to super-pixels and then segments them.


\section{Medieval Manuscripts}
\label{toc:experimental_setting}

\global\csname @topnum\endcsname 0
\global\csname @botnum\endcsname 0

In the last years the \ac{DIA} field has witnessed significant improvements in terms of layout analysis performance~\cite{antonacopoulos2007special,namboodiri2007document,alberti2017lda}.
This led researchers to focus on more challenging data, such as historical documents.
Historical documents have a complex layout, high variance across samples and present several sources of noise in the form of colour and degradation.
Thus, algorithms need to be very robust and flexible, making this kind of data a natural candidate for benchmarking novel ideas and techniques.

\subsection{DIVA-HisDB}
\label{toc:sub:diva_hisdb}

In this work, we mainly use the DIVA-HisDB dataset, which is a collection of three medieval manuscripts that have been selected for the high complexity of their layout \cite{simistira2017icdar2017}. 
Figure \ref{fig:dataset} shows some properties of the dataset that make it particularly challenging.
The complexity of the layout is increased by marginal and/or interlinear glosses, additions, and corrections (c, d).
Moreover, these elements often have only a small difference in colour and style with the main text (b),
Decorations are one of the main challenges as they come in different colours (d, e) and sizes.
There are lettrines (the large initial letters often used in the beginning of a new paragraph) depicted with red marks (b) and with radically different sizes (c), which makes it difficult to distinguish text from decorations. 
The main text is written with different scripts and languages (across manuscripts), but most importantly, with different script sizes (d, e) and colours (c, d).
Finally, some paper degradation can make the text hard to read (a) or even impossible (f).
These kinds of artifacts would pose a significant challenge to traditional binarization methods. 

The dataset consists of 150~pages, and it is publicly available\footnote{\url{http://diuf.unifr.ch/hisdoc/diva-hisdb}}. 
The RGB images are in JPG format, captured with a camera at 600 dpi for an average of \textasciitilde 19 megapixels per image.
The ground truth is available both at pixel and text-line level.
The pixel labelling is done with four different annotated classes: main-text-body, decorations, comments, and background.
These classes might overlap, making the task harder as it is multi-class and multi-label.
The text-line annotations are tight polygons corresponding for each text-line saved into a PAGE XML~\cite{pletschacher2010page} file.

\section{Task: Text-Line Segmentation}
\label{toc:task}

\begin{figure}[t]
    \includegraphics[width=\columnwidth]{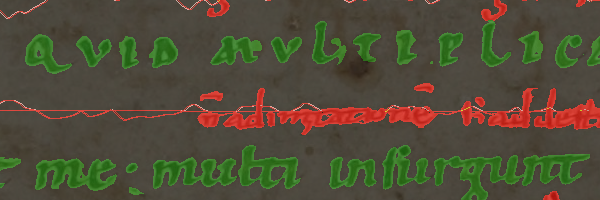}
    \caption{%
    Example of Wavelength \cite{seuret2017wavelength} seam-carving based algorithm which does not produce tight polygons around text-lines.
    Green denotes the foreground pixels correctly segmented into text-lines.
    Red denotes foreground pixels which should not have been part of a text-line, e.g. glosses or decorations.
    Note how the interlinear glosses, in this case, are enclosed into the polygons of two different text-lines and hence marked in red.
    This hinders the quality of the text-line extraction process, thus supporting our choice to produce tight polygons around the text. 
    } 
    \label{fig:tight_polygons}
\end{figure}

The task considered in this work is text-line segmentation i.e. given an RGB image (Figure~\ref{fig:task_rgb}), produce a list of polygons enclosing the text-lines in the document (Figure~\ref{fig:task_lines}); as proposed for Task-3 at the ICDAR2017 Competition on Layout Analysis for Challenging Medieval Manuscripts \cite{simistira2017icdar2017}.
These polygons should be distinct, non-overlapping and non-convex to be effective on the DIVA-HisDB dataset.
Simple rectangular bounding boxes or loose polygons are not well suited to deal with the interlinear glosses.
Figure~\ref{fig:tight_polygons} shows a visualised output of a seam carving method which does not produce tight polygons and consequently fails to reject interlinear glosses as being part for text-line, resulting in a lower text-line extraction score. 
Producing tight polygons is harder than estimating the baseline of the text-line, which is a very different task often found in the literature.

\subsection{Evaluation of the Task}
\label{toc:sub:task_evaluation}

To evaluate the performance of our system we follow the protocol initially proposed in the competition. 
We compute two different scores.
The main one measures how well are the text-line being segmented, i.e. if all lines have are extracted correctly and there are no extra lines.
The second one measures whether all the pixels belonging to text-lines have are enclosed into polygons - without extra content such as glosses or decorations.
Both of them are computed with the Intersection over Union (IU) metric, which is a statistic used for comparing the similarity and diversity of sets \cite{levandowsky1971}. 
Unlike accuracy, this metric is invariant to the total number of samples, and it is more strict than the F1-score when a mistake occurs, as the true positive cases are not weighted twice \cite{alberti2017evaluation}.
To compute these scores and produce the visualisations shown in this paper we use the open-source\footnote{\url{https://github.com/DIVA-DIA/DIVA_Line_Segmentation_Evaluator}} competition tool.

\section{Method Description}
\label{toc:method}

\begin{figure}[!t]
\centering

\subfloat[Original RGB domain]{
\includegraphics[width=\columnwidth, height=2.2875cm]{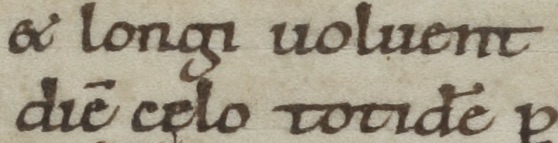}\label{fig:step_1}}

\subfloat[Pixel-labelled domain]{
\includegraphics[width=\columnwidth, height=2.2875cm]{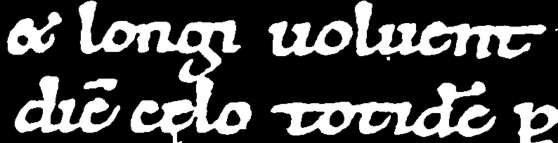}\label{fig:step_2}}

\subfloat[Energy map]{
\includegraphics[width=\columnwidth, height=2.2875cm]{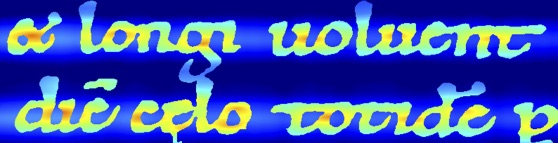}\label{fig:step_3}}

\subfloat[Seams cutting the energy map]{
\includegraphics[width=\columnwidth, height=2.2875cm]{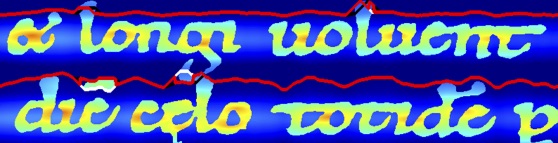}\label{fig:step_4}}

\subfloat[Binning algorithm]{
\includegraphics[width=\columnwidth, height=2.2875cm]{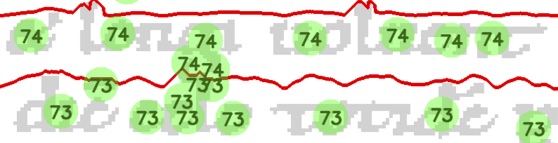}\label{fig:step_5}}

\subfloat[Canvas with drawn MST and CC]{
\includegraphics[width=\columnwidth, height=2.2875cm]{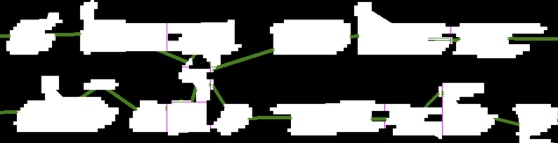}\label{fig:step_6}}

\subfloat[Output polygons overlapped with RGB image]{
\includegraphics[width=\columnwidth, height=2.2875cm]{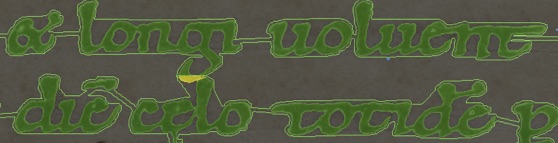}\label{fig:step_7}}

\caption{%
Visualisation of the steps of our proposed method.
Start from the RGB domain (a) we perform semantic segmentation at pixel-level (b).
We compute a custom energy map (c) on which we use a simple seam carving algorithm (d).
We then bin each centroid according to their location in respect of the seams (e) and finally compute the enclosing polygon around them (f).
The output can then be overlapped on the RGB domain visualisation purposes (g).
Section~\ref{toc:method} describes each step in detail.
} 
\label{fig:method_steps}
\end{figure}

In this section, we provide a detailed explanation of our text-line segmentation method.
The key intuition lies behind splitting the task into two separate steps and to solve them separately. 
First, we perform semantic segmentation at the pixel level, i.e. for each pixel, we assign one or more labels from: main-text-body, decorations, comments, and background.
Second, we extract the text-lines with a novel algorithm which is designed to leverage the high-quality semantic segmentation to deliver precise and tight polygons around the text-lines. 
The rest of this section describes each step of the algorithm in more detail. 

\subsection{Semantic Segmentation}
\label{toc:sub:method:step_2}

In order to process the input from RGB domain (Figure~\ref{fig:step_1}) into a pixel-labelled domain (Figure~\ref{fig:step_2}), we perform semantic segmentation at pixel level using the vanilla ResNet-18 \cite{he2015resnet}, which delivers near state-of-the-art results \cite{pondenkandath2017exploiting, simistira2017icdar2017}.
To run it we used the DeepDIVA deep-learning framework \cite{albertipondenkandath2018deepdiva}.
The quality of the semantic segmentation directly correlates with the quality of the final text-line extraction.
However, since existing - and reproducible - solutions provide excellent results on this task, we did not need to invest additional resources in improving this step of the work-flow. 

To polish the output of the network, we apply simple denoising techniques, such as removing small components and the like. 
At this point, we have an image in the labels domain, where we can select only the main text pixels and discard everything else - such as glosses and decorations - thus producing a clean main text filtered image (Figure~\ref{fig:step_2}).

\subsection{Building the Energy Map}
\label{toc:sub:method:step_3}

Preparing the energy map is a critical step for the success of seam-carving based algorithms.
A vital contribution of this paper is the intuition to start building the energy map from the pixel-labelled domain and not from the raw RGB domain.
This ensures that most sources of noise have been filtered out, resulting in a clean starting point.

Let's define a binary filtered text image $x = (x_1, .., x_{n\times m}) \in \{ 0, 1 \}^{n\times m}$ in the pixel domain where the value of $1$ is assigned in correspondence of a pixel classified as text  (Figure~\ref{fig:step_2}). 
Our energy map $E = ( e_1, \dots, e_{n \times m})$ (Figure~\ref{fig:step_3}) is built by summing three components that we call: the background energy $B = ( b_1, \dots, b_{n \times m})$, the text energy $T = ( t_1, \dots, t_{n \times m})$, and the smoothed energy $S = ( s_1, \dots, s_{n \times m})$. Formally, computed as below:
\begin{equation}
\label{eq:E}
    E(x) = B(x) + T(B(x)) + S(B(x), T(B(x)))
\end{equation}

\subsubsection{Background Energy}
First, we find the centroid of all the \acp{CC} in the image. 
For each pixel we then compute the distance to the closest \ac{CC} using Euclidean distance, producing a distance map.
The background energy $B(x)$ is the inverse of the distance map (Equation~\ref{eq:B}).
In this resulting map, the closer is a pixel to a centroid the higher is its energy. 
Conversely, if a pixel is far away from the text is has deficient energy.

\begin{equation}
\label{eq:B}
    B_i(x_i) =   \frac{1}{ \min\limits_{c ~ \in ~ CC} \lVert l(x_i) - l(c) \rVert } \quad \forall i = 1, \dots, n \times m
\end{equation}

where $l(\cdot)$ resolves the coordinates of that pixel in the image.

\subsubsection{Text Energy}
The text energy is a copy of the background energy where the value of pixels which do not belong to the main text - according to the semantic segmentation output - is set to zero (Equation~\ref{eq:T}).
This is done to raise the importance of text pixels.
In fact, after merging this map with the background energy, pixels belonging to text have twice the energy of pixels which are at the same distance to a centroid but not on the text.

\begin{equation}
\label{eq:T}
    T_i(B_i(x_i)) =  \begin{cases}
                    B_i(x_i)        &  x_i = 1\\
                    0               &  x_i = 0
                    \end{cases}  \quad \forall i = 1, \dots, n \times m
\end{equation}

\subsubsection{Smoothed Energy}
The third component, referred to as the smoothed energy, is necessary to both close the gaps between words and to remove high-frequency noise patterns.
We compute the sum of text $T(B(x))$ and background $B(x)$ energies and then perform convolution with two different kernels sequentially: first with a global smoothing kernel $k_1$ then with a local one  $k_2$ (Equation~\ref{eq:S}).
The global kernel $k_1$ is of size $n \times n$ and has the shape of a centred ``+'' sign of ones spanning the entire kernel with zeroes everywhere else.
This provides a noisy but global estimation of the energy distribution over the other two energy components (background and text energies).
The local kernel $k_2$ is a regular $32 \times 32$ averaging filter which purpose is to reduce the amplitude of the high-frequency noise produced by the global averaging.
Note that this entire process is symmetric in the vertical and horizontal axis, thus entirely agnostic to the text orientation as opposed by the often used method of projection profiles signals.
Moreover, we would like to stress that these smoothing convolutions are performed on the sum of text and background energies and not on the semantic segmentation domain.

\begin{equation}
\label{eq:S}
    S(B(x), T(B(x))) = C_2(C_1(T(B(x)) + B(x)))
\end{equation}

where $C_1$ and $C_2$ denote the convolution operations with the kernels $k_1$ and $k_2$ respectively.

\subsection{Casting the Seams}
\label{toc:sub:method:step_4}

Once the energy map is computed, we cast a seam every $\alpha$ vertical pixels from both left to right (black line), and right to left (white line) as shown in Figure~\ref{fig:step_4}. 
We use the standard seam-carving algorithm with the addition of a penalty term $\beta$ for deviating from the horizontal axis.
To reduce the noise, whenever any two seams intersect twice (visually forming an ellipse on the image), we keep the fittest seam (red line) and merge the other onto it (for that region only).
The fitness of a seam is the inverse of the energy accumulated over the pixels it traverses.

The number of pixels between the starting position of two seams $\alpha$ and the penalty term $\beta$, are the only parameters in our text-line extraction method.
However, as described in Section~\ref{toc:parameter_robustness}, the results are fairly stable in respect to them.
This is a desired property of our method, which strives to be parameter-free. 

\subsection{Binning the Centroids}
\label{toc:sub:method:step_5}

The next step in the pipeline uses the final seams location to cluster the centroids together, where each cluster represents a line. 
To do so, we first count for each centroid how many seams are below it (Figure~\ref{fig:step_5}).
This allows us to put each centroid into different bins (clusters) based on this value.
The intuition is that if the seams do not cross a text-line, all centroids belonging to it have the same amount of seams below them.
Occasionally we observed that a specific bin contains only a few centroids (typically one or two). 
This happens when a ``i'' dot or the upper part of the letter ``T'' get separated by the body of the character by one or more seams. 
In this case, we merge such small bins to the closest bin nearby using Euclidean distance.

\subsection{Extracting the Polygons}
\label{toc:sub:method:step_6}

Once we have computed which centroids belong to which lines, we only need to compute the enclosing polygon around them.
Given a set of centroids from the previous step, we draw their corresponding \ac{CC} and the \ac{MST} connecting their centroids on an empty canvas.
We then convolve this canvas with a small $5\times 5$ averaging kernel such that the final result is a blurred version of the text-line (Figure~\ref{fig:step_6}).
At this point we extract the - only - connected component existing on the canvas and its contour points are our enclosing polygon.

\section{Results}
\label{toc:results}

\begin{table}[t] 
    \caption{%
    Results of text-line extraction on the DIVA-HisDB dataset (see Section~\ref{toc:sub:diva_hisdb} measured with the competition tool(see Section~\ref{toc:sub:task_evaluation}.
    Our proposed method outperforms state-of-the-art results by reducing the error by 80.7\% and achieving nearly perfect results.
    Methods with * notation use semantic segmentation at pixel-level as pre-processing step.
    }
    \label{tab:results}
    \begin{center}
    \begin{small}
    \begin{sc}
    \resizebox{0.99\columnwidth}{!}{%
    \begin{tabular}{l c c }
    \toprule
    Method & Line iu \% & Pixel iu \% \\
    \midrule
    wavelength \cite{seuret2017wavelength}                               & 68.58 & 79.13 \\
    Brigham Young University \cite{simistira2017icdar2017}                     & 81.50 & 83.07 \\
    CITlab Argus LineDetect \cite{gruuening2017robust}                   & 96.99 & 93.01 \\
    wavelength* (tight polygons) \cite{simistira2017icdar2017}           & 97.86 & 97.05 \\
    proposed method*                                                     & \textbf{99.42} & 96.11 \\
    \bottomrule
    \end{tabular}}
    \end{sc}
    \end{small}
    \end{center}
\end{table}

\begin{table}[t] 
    \caption{%
    Results of the experiments shown in Table~\ref{tab:results} with the difference that every method listed has received the ground truth of the semantic segmentation at pixel-level as input.
    Our proposed text-line extraction method is superior to state-of-the-art even if both methods run on the same perfect input.
    Moreover, in our experience, an algorithm which is not designed to take advantage of this pre-processing step will not benefit from it.
    }
    \label{tab:results_from_gt}
    \begin{center}
    \begin{small}
    \begin{sc}
    \resizebox{0.99\columnwidth}{!}{%
    \begin{tabular}{l c c }
    \toprule
    Method from gt & Line iu \% & Pixel iu \% \\
    \midrule
    wavelength \cite{seuret2017wavelength}                      & 66.44          & 81.52    \\
    wavelength (tight polygons) \cite{simistira2017icdar2017}   & 99.25          & 98.95    \\
    proposed method                                             & \textbf{100.0} & 97.22     \\
    \bottomrule
    \end{tabular}}
    \end{sc}
    \end{small}
    \end{center}
    \vspace{-5mm}
\end{table}

\begin{figure}[t]
    \centering
    \includegraphics[width=\columnwidth]{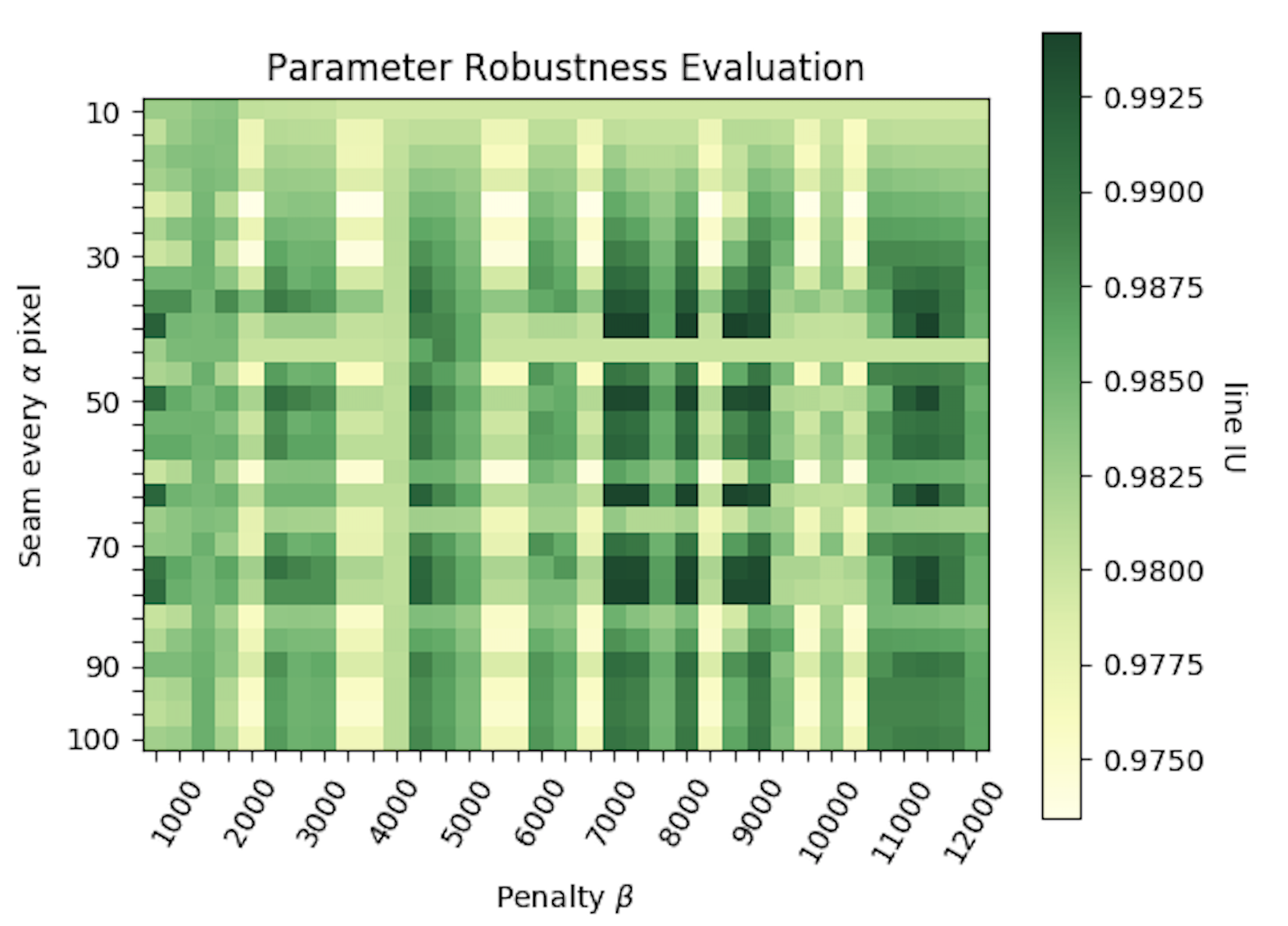}
    \vspace{-5mm}
    \caption{%
    In this heatmap, we show the robustness of our text-line extraction method to two parameters: the number of pixels between the starting position of two seams $\alpha$ and penalty term $\beta$.
    Note that the lower end of the heatmap scale compares favourably with state-of-the-art (see Table~\ref{tab:results}) meaning that regardless of the choice of parameters, our method produces excellent results.
    Because of this, it is in practice virtually a parameter-free method.
    } 
    \label{fig:heatmap}
    \vspace{-3mm}
\end{figure}

\begin{figure}[!t]
    \centering
    
    \subfloat[HDRC-Chinese dataset]{
    \includegraphics[width=.47\columnwidth]{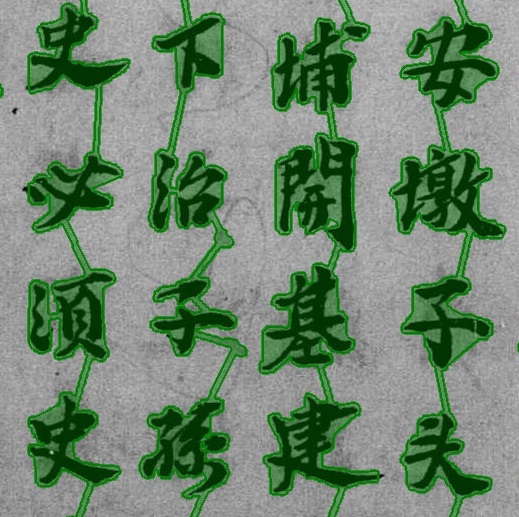}}
    \hfil
    \subfloat[ICDAR 2013 Contest]{
    \includegraphics[width=.47\columnwidth]{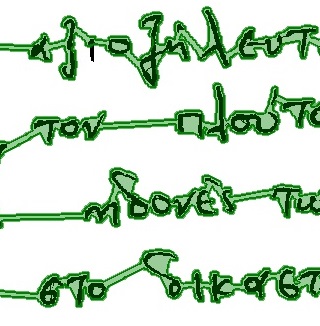}}
    \vfil
    \subfloat[WAHAD Database]{
    \includegraphics[width=.47\columnwidth]{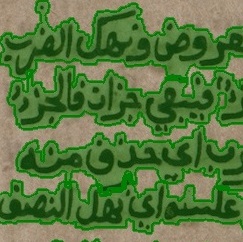}}
    \hfil
    \subfloat[George Washinton]{
    \includegraphics[width=.47\columnwidth]{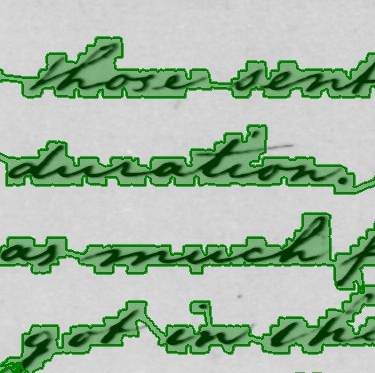}}
    
    \caption{
    In this figure, we show the visualisation of preliminary experiments on other datasets.
    We show that our method performs reasonably well on ICDAR 2019 HDRC-Chinese dataset (a), ICDAR 2013 Handwriting Segmentation Contest dataset (b), Challenging Handwritten Dataset from the WAHAD Database (c) and the well known George Washiton dataset (d) too. 
    } 
    \label{fig:other_datasets}
    \vspace{-3mm}
\end{figure}

In this section, we present a comprehensive evaluation of the performances of our proposed method.

In Table~\ref{tab:results} there are the results on the DIVA-HisDB dataset (see Section~\ref{toc:sub:diva_hisdb}).
Our method achieves nearly perfect results (99.42\%) and outperforms state-of-the-art (97.86\%) resulting in a error reduction of 80.7\%. 
We investigated the mistakes of our algorithm and found that only 4 of the 871 lines have been mis-detected.
The reason in all cases is a fault in the semantic segmentation step, where a decoration (which typically pose the hardest challenge in this dataset) has been not correctly classified, resulting in a ghost/extra line at the end of the pipeline. 
This calls for a higher quality semantic segmentation.

We asked ourselves ``what if'' we had a perfect pixel-labelled segmentation tool? How good could we extract the text-lines?
The answer is in Table~\ref{tab:results_from_gt} where we performed the same task, but this time we swapped our semantic segmentation network with the pixel-level ground-truth provided along with the data.
This represents the upper-bound performances, as no tool will produce a better segmentation than the ground-truth.
In this scenario our method performed at 100\% line IU, reinforcing our previous observation that our text-line extraction method has made the mistakes only in the presence of wrong results from the semantic segmentation step. 
We critically questioned if this perfect score was not an artefact of running from the ground truth, i.e. would any other algorithm perform so well if given the chance to start from that point?
The answer is no.
Two conclusions can be made from these experiments.
The first is that our proposed text-line extraction method is superior to state-of-the-art even if both methods are given the same perfect semantic segmentation input.
The second is that, in our experience, an algorithm which is not designed to take advantage of this pre-processing step does not benefit from it - which is unsurprising to a certain extent.

\subsection{Parameter Robustness}
\label{toc:parameter_robustness}

Even an exemplary algorithm becomes less appealing if it requires extensive parameter optimisation for it to work. 
Therefore our goal was to design a parameter-free algorithm. 
We did not succeed at that and ended up with two parameters on our hands: the number of pixels between starting position of two seams and penalty term.
However, we observed that the results were stable with respect to the values chosen for them.
We then measured this empirically, and the results are visualised in the heatmap in Figure~\ref{fig:heatmap}.
Note that the lower end of the heatmap scale compares favourably with state-of-the-art (see Table~\ref{tab:results}) meaning that regardless of the choice of parameters, our method produces excellent results.
Because of this, it is in practice virtually a parameter-free method.
 
\subsection{Extension to other datasets}

To verify if the performances of our algorithm generalise to other scenarios, we ran it on other datasets.
Specifically, we choose the ICDAR 2019 HDRC-Chinese dataset~\cite{simistira2019icdar2019}, the ICDAR 2013 Handwriting Segmentation Contest dataset~\cite{stamatopoulos2013icdar}, the challenging handwritten dataset from the WAHAD Database~\cite{abdelhaleem2017wahd} and the well known George Washingtion dataset~\cite{fischer2012lexicon}.
Preliminary results visualised in Figure~\ref{fig:other_datasets} suggest our proposed method is performing appreciably well on other datasets too.
Despite the promising preliminary results, additional experiments would be required to assess the generalisation properties of the algorithm.
Note that if one desires to push the performance on a dataset specifically is it always possible to develop ad-hoc strategies for it, but this was not the case for this paper as we wanted to propose a general methodology rather than a particular algorithm.

\section{Limitations: No Free Lunch Theorem}
\label{toc:limitations}

In this section, we discuss some known limitations of our method.
In fact, despite the encouraging results presented, there is still a long way before having an algorithm which can successfully and reliably tackle the general case of any historical document.
Specifically, we make some assumptions on the type of documents we tackle.
The first is that the text has the same orientation within the document, i.e. documents with mixed vertical/horizontal text or extremely bent lines would not be processed successfully.
The reason is the penalty term on the seam-carving phase which forces the seams to follow a given direction: either horizontal or vertical.
Second, in this work, we don't leverage structural layout of the page, thus, if the layout is more complicated than single/double column format we would need a layout analysis pre-processing step to isolate the different text blocks.
An example of such a document is a grid-like pattern document with different amount of lines in each cell, e.g. in the ICDAR 2019 HDRC-Chinese dataset. 
Third, our methods require a excellent pixel-level semantic segmentation.
Despite there are open-source and well documented state-of-the-art solutions, this step is computationally intensive and requires a non-negligible amount of training data with ground truth at pixel-level.
Therefore, the difficulty of that step must not be underestimated.

\section{Conclusion}
\label{toc:conclusion}

In this paper, we tackle the task of text-line extraction in the context of historical document image analysis.
We show how semantic segmentation at pixel-level can be used as strong denoising pre-processing step before performing text-line extraction.
Evidence suggests that the higher the quality of semantic segmentation the better the results of the text-line extraction step. 
This calls for further research on developing a stable and reliable semantic segmentation technology which can work on different types of documents.
We also introduce a novel and robust algorithm that leverages this data preparation achieving nearly perfect results on a challenging dataset of medieval manuscripts.
Preliminary experiments showed promising results on other datasets, but a thorough investigation is still required to quantitatively asses the performances of the algorithm in a general case.
With this work, we take a step further towards improving the automated processing of historical documents.

\section*{Acknowledgment}
The work presented in this paper has been partially supported by the HisDoc III project funded by the Swiss National Science Foundation with the grant number $205120$\textunderscore$169618$.

\bibliography{biblio}
\bibliographystyle{IEEEtran}

\end{document}